\documentclass[10pt,twocolumn,letterpaper]{article}

\usepackage{iccv}
\usepackage{times}
\usepackage{epsfig}
\usepackage{graphicx}
\usepackage{amsmath}
\usepackage{amssymb}
\usepackage{multirow}
\usepackage{booktabs}
\usepackage{bbding}
\usepackage[misc]{ifsym}

\usepackage[breaklinks=true,bookmarks=false]{hyperref}

\iccvfinalcopy 

\def\iccvPaperID{****} 

\ificcvfinal\fi

\begin{document}

\title{AttT2M: Text-Driven Human Motion Generation with Multi-Perspective Attention Mechanism}

\author{Chongyang Zhong\textsuperscript{1,2}\thanks{Equal contribution}, Lei Hu\textsuperscript{1,2}\footnotemark[1], Zihao Zhang\textsuperscript{1}\footnotemark[1], Shihong Xia\textsuperscript{1,2}\thanks{Corresponding author}\\ 
$^1$Institute of Computing Technology, Chinese Academy of Sciences\\
$^2$University of Chinese Academy of Sciences\\
{\tt\small \{zhongchongyang, hulei19z, zhangzihao, xsh\}@ict.ac.cn}
}

\maketitle
\ificcvfinal\thispagestyle{empty}\fi

\begin{abstract}
Generating 3D human motion based on textual descriptions has been a research focus in recent years. It requires the generated motion to be diverse, natural, and conform to the textual description. Due to the complex spatio-temporal nature of human motion and the difficulty in learning the cross-modal relationship between text and motion, text-driven motion generation is still a challenging problem. To address these issues, we propose \textbf{AttT2M}, a two-stage method with multi-perspective attention mechanism: \textbf{body-part attention} and \textbf{global-local motion-text attention}. The former focuses on the motion embedding perspective, which means introducing a body-part spatio-temporal encoder into VQ-VAE to learn a more expressive discrete latent space. The latter is from the cross-modal perspective, which is used to learn the sentence-level and word-level motion-text cross-modal relationship. The text-driven motion is finally generated with a generative transformer. Extensive experiments conducted on HumanML3D and KIT-ML demonstrate that our method outperforms the current state-of-the-art works in terms of qualitative and quantitative evaluation, and achieve fine-grained synthesis and action2motion. Our code is in \href{https://github.com/ZcyMonkey/AttT2M}{https://github.com/ZcyMonkey/AttT2M}.
\end{abstract}

\section{Introduction}
\label{sec:intro}

\begin{figure}[t]
    \centering
    \includegraphics[width = 0.47\textwidth,height = 10cm]{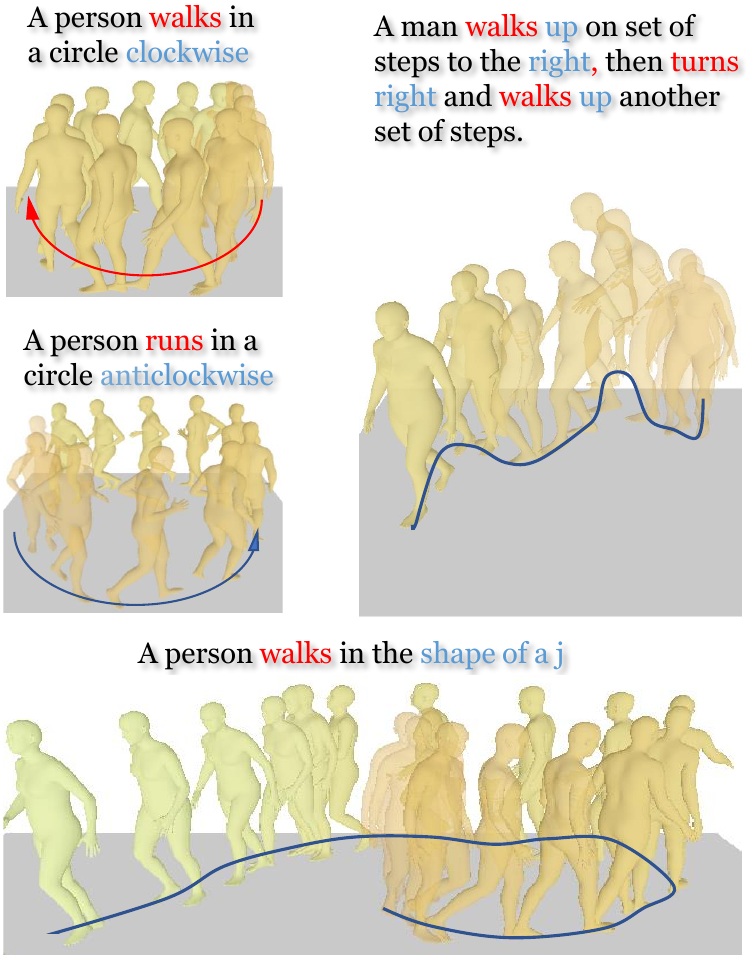}
    \caption{\textbf{Text-driven generation visualization.} Our method achieves high-quality motion that is precisely consistent with semantically rich text descriptions. }
    \label{img:01}
    \vspace{-10pt}
\end{figure}

With the expansion of application scenarios and demands, cross-modal human motion synthesis has become a hot research topic in computer vision and graphics. Among them, text-driven human motion synthesis aims to synthesize natural human motion that matches a given text description, which can be applied to intelligent animation production, virtual reality, game and film industry, human-robotics interaction, etc.

However, introducing natural language descriptions as constraints for motion synthesis, which means using cross-modal high-level semantics to control the motion synthesis process, is a difficult interdisciplinary problem involving natural language processing, motion modeling, and cross-modal relationship learning. With the availability of pre-trained language models such as CLIP~\cite{radford2021learning}, BERT~\cite{devlin2018bert}, and GPT~\cite{radford2018improving} for natural language processing, there are still two challenges that remained: (1) human motion is high-dimensional data with irregular spatial structure and highly dynamic temporal characteristics. It is challenging to perform text-driven synthesis in pose space directly; (2) there are complex local (specific words correspond to specific motion sub-sequences) and global (the overall text description corresponds to the order and connection of these sub-sequences) correspondence between text and motion. How to better learn this cross-modal relationship is still a problem to be solved.

Compared to some works that directly generated text-driven motion in the pose space\cite{ahuja2019language2pose,ahn2018text2action,bhattacharya2021text2gestures,zhang2022motiondiffuse,tevet2023human}, others attempted to learn the low-dimensional representation of motion first using auto-encoder~\cite{guo2022generating}, VAE~\cite{chen2023mld} and VQ-VAE~\cite{zhang2023t2m} with temporal encoding. However, since human motion has both spatial and temporal characteristics, it is necessary to consider both when dealing with latent representation learning~\cite{zhong2022spatio,yan2018spatial}. Therefore, we propose Vector Quantised-Variational AutoEncoder(VQ-VAE) with \textbf{Body-Part attention-based Spatio-Temporal(BPST)} encoder to learn an expressive latent space. As for the motion-text cross-modal relationship, some works used cross-modal translation to learn the shared representation of text and motion and directly generated motion from text feature~\cite{ahuja2019language2pose,ahn2018text2action,petrovich2022temos,athanasiou2022teach}. However, the vast gaps between motion and text data make learning an adequate shared representation difficult. Other researchers attempted to treat text information as a condition during the motion generation~\cite{guo2022generating,zhang2023t2m,chen2023mld,zhang2022motiondiffuse,tevet2023human,guo2022tm2t}. We observe two levels of correspondence in the cross-modal relationship between motion and text: 1). the local correspondence between words of text and sub-segments of motion sequences; 2). the global correspondence between the overall semantics of text and the whole motion sequence. Previous research typically focused on either local~\cite{zhang2022motiondiffuse} or global~\cite{zhang2023t2m, tevet2023human} text information or combined them by simply concatenating~\cite{guo2022generating}. We propose \textbf{Global and Local Attention(GLA)} to consider both levels more reasonably better to learn the cross-modal relationship between text and motion.

\begin{figure*}[t]
    \centering
    \includegraphics[width = 1\textwidth]{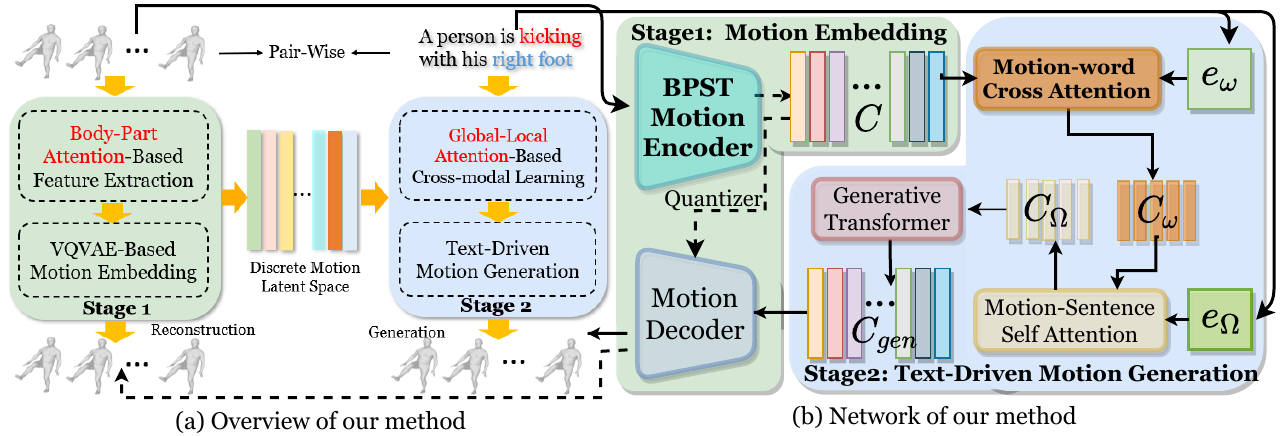}
    \caption{\textbf{Overview and network}. (a)We propose a text-driven motion generation method based on a multi-perspective attention mechanism, including body-part attention for expressive spatio-temporal feature extraction in the motion embedding stage and global-local attention for a better cross-modal relationship learning in the text-driven generation stage. (b)In stage 1, we use body-part attention-based spatio-temporal VQ-VAE to learn a discrete latent motion representation. In stage 2, we calculate local motion-word cross-attention and global motion-sentence conditional self-attention to ensure that the motion generated by the generative transformer is highly consistent with text descriptions.}
    \label{img:02}
\end{figure*}

Specifically, we propose AttT2M, a two-stage text-driven motion generation model with multi-perspective attention(Figure~\ref{img:02}(a)). The first attention focuses on motion embedding perspective. We use a spatial transformer~\cite{vaswani2017attention} based on \textbf{body-part attention} and TCN to extract the spatio-temporal features, and then a VQ-VAE~\cite{van2017neural} is used to quantize the features into a discrete codebook. In the text-diven generation stage, we propose \textbf{global and local attention} from the perspective of cross-modal relationship learning. After the global (sentence-level) and local (word-level) text features are extracted using CLIP, we calculate the motion-word cross-attention to learn the local cross-modal relationship and learn the global cross-modal relationship by motion-sentence conditional self-attention. Finally, a generative transformer is used to generate motion sequences.

We conduct extensive qualitative and quantitative experiments on two widely used datasets, KIT-ML~\cite{plappert2016kit} and HumanML3D~\cite{guo2022generating}. The experimental results show that our work achieves better text-driven motion generation results than previous work in qualitative and quantitative comparisons. The generated motions highly match the given natural language descriptions and maintain reliable diversity while being highly realistic and natural(seeing Figure~\ref{img:01}). Our work can also achieve fine-grained generation and action to motion.

Our contributions can be summarized as follows:
\vspace{-5pt}
\begin{itemize}
\item 1. We propose a Body-Part attention-based Spatio-Temporal VQ-VAE to map motion sequences into a better discrete code book, resulting in a more expressive low-dimensional motion representation.
\item 2. We introduce Global and Local Attention to learn the global-local cross-modal relationship between text and motion, achieving precise correspondences between motion and text.
\item 3. We conduct extensive qualitative and quantitative experiments on KIT-ML and HumanML3D to demonstrate that our method outperforms the previous state-of-the-art methods, which can also handle fine-grained and action-to-motion generation.
\end{itemize}

\section{Related Works}

\textbf{Human motion synthesis}
is a long-standing research topic~\cite{xia2017survey}. Here we only introduce works based on deep learning and exclude earlier methods. Motion synthesis started with the unconstrained generation, which generated realistic and natural motion sequences in completely random ways~\cite{fragkiadaki2015recurrent,li2017auto,zhang2020perpetual}. With the development of research and increasing demands, methods based on different conditions have been proposed. One primary condition was motion content. Some works focused on generating motion that satisfied the constraints of low-level control signals (such as speed, direction, trajectory, etc.). They either used a two-stage approach\cite{holden2016deep,wang2019combining,ling2020character} by training a generative model as a prior and then generated motions through optimization, or used an end-to-end manner by parameterizing the control signals as input to predict the control signals and pose for the next frame~\cite{lee2018interactive,holden2017phase,zhang2018mode,starke2019neural,starke2020local,starke2021neural}. Other researchers attempted to use motion graph to synthesize high-quality human motion through searching and blending~\cite{clavet2016motion,
holden2020learned}. In addition, another content constraint was the past motion sequences(referred to as diverse motion prediction). They usually used generative models such as VAE~\cite{walker2017pose,aliakbarian2020stochastic} and GAN~\cite{barsoum2018hp,kundu2019bihmp} to encode past motion sequences and then sampled in the latent space to generate multiple prediction results. Recently, motion generation based on the cross-modal condition has been a hot topic. Some works hoped to generate dances from music~\cite{lee2019dancing,li2020learning,valle2021transflower}, while other researchers were trying to generate motion from the given text, which included action labels~\cite{petrovich2021action,guo2020action2motion,zhong2022learning} as well as natural language descriptions~\cite{ghosh2021synthesis,ahuja2019language2pose,bhattacharya2021text2gestures,ahn2018text2action,Lin2018GeneratingAV,petrovich2022temos,athanasiou2022teach,guo2022generating,zhang2023t2m,chen2023mld,zhang2022motiondiffuse,tevet2023human,guo2022tm2t,tevet2022motionclip,hong2022avatarclip}.

\textbf{Motion latent representation learning}
in motion synthesis is about mapping motions into a low-dimensional latent space due to the high-dimensional and complex nature of the original pose space. Some works utilized autoencoder~\cite{guo2022generating} structures to achieve dimension reduction. Other works attempted to train a generative model based on VAE~\cite{chen2023mld} or VQ-VAE~\cite{zhang2023t2m} to learn a latent representation of motion with more diversity. Most of these works only focused on temporal characteristics, using RNNs, TCNs, or transformers to learn temporal features. However, in motion prediction and action recognition, researchers had found that the spatial features were also crucial in motion latent representation learning~\cite{zhong2022spatio,yan2018spatial}. So we propose Body-Part attention-based SpatiO-Temporal VQ-VAE to learn a more expressive latent representation.

\textbf{Text-driven motion generation}
aims to generate 3D human motion based on textual descriptions. We categorize related works into three types. The first type desired to learn shared latent representations of motion and text, and directly converted text to motion. After Text2Action~\cite{ahn2018text2action} and Language2pose~\cite{ahuja2019language2pose} made the first attempts, TEMOS~\cite{petrovich2022temos} introduced a transformer VAE architecture to make the latent distributions of text and motion as similar as possible, and TEACH~\cite{athanasiou2022teach} achieved coherent motion generation based on multiple action labels. The second type rendered 3D human bodies as images and learned the motion-image-text relationship using contrastive learning with the help of CLIP. MotionCLIP~\cite{tevet2022motionclip} focused on motion generation, while AvatarCLIP~\cite{hong2022avatarclip} generated human appearance and simple motion simultaneously.  The third type treated text as a condition for motion generation. Some works were based on an encoder-decoder framework, using transformers~\cite{bhattacharya2021text2gestures} or RNNs ~\cite{Lin2018GeneratingAV} to encode motion and concatenating it with text feature as input to the decoder. To increase the diversity of generated motion, generative models were introduced. RNN+CVAE~\cite{guo2020action2motion}, transformer+CVAE~\cite{petrovich2021action}, and Uncoupled-modulation CVAE~\cite{zhong2022learning} are used to generate motion from action labels. With the proposal of a large-scale motion-language dataset called HumanML3D~\cite{guo2022generating}, many studies achieved better results. T2M~\cite{guo2022generating} adopted a two-stage approach, learning motion clip codes and conducting text-driven motion generation. They also achieved better results by jointly training text-to-motion and motion-to-text tasks~\cite{guo2022tm2t}. T2M-GPT~\cite{zhang2023t2m} proposed to learn the discrete representation with VQ-VAE and used a generative transformer for generation. Other works attempted to use the diffusion model~\cite{ho2020denoising} in pose space directly~\cite{zhang2022motiondiffuse,tevet2023human} or latent space~\cite{chen2023mld} for text-driven motion generation. This paper proposes a Global-Local Attention mechanism to learn multi-level cross-modal relationships between motion and text, achieving better text-driven motion generation.

\section{Our Method}
The problem we aim to solve is text-driven motion generation, which refers to generating a T frames sequence of human motions $X=\{x_1,x_2,...,x_T\}$ based on a given text description $\Omega=\{\Omega_1,\Omega_1,...,\Omega_N\}$ consisting of $N$ words.

\subsection{Method Overview}
To generate high-quality and text-corresponding motion, we explore motion latent space learning and text-motion cross-modal relationship learning. The proposed two-stage motion generation framework is shown in Figure~\ref{img:02}(b).
Stage 1 is motion embedding. We rearrange the original motion representation to adapt the body-part segmentation and use a Body-Part attention-based Spatio-Temporal(BPST) motion encoder to learn an expressive spatio-temporal feature. Then a VQ-VAE is used to learn a discrete motion latent space $C$ through quantizer. In Stage 2, we learn the cross-modal relationship between text and motion on the global and local levels to generate motions according to given text $\Omega$. After the word-level embedding $e_\omega$ and sentence-level embedding $e_\Omega$ are extracted by CLIP, we propose motion-word cross attention to learn the local cross-modal relationship. Moreover, we compute motion-sentence self-attention to learn the global cross-modal relationship. To generate diverse results, we use a generative transformer to generate motion code $C_{gen}$, and the final generated sequence is obtained by the motion decoder.

\subsection{The Motion Embedding Module}
The pose of motion exhibits infinite variations, leading to a high-dimensional pose space that makes it difficult to model human motions. Therefore, many researchers attempt to learn a latent space to reduce the dimension~\cite{guo2022generating,chen2023mld,zhang2023t2m}. However, 3D human motion is a structured time series with spatial and temporal characteristics. Previous text-driven motion generation works paid more attention to temporal modeling using RNN~\cite{guo2020action2motion}, TCN~\cite{guo2022generating,zhang2023t2m}, Transformer~\cite{petrovich2022temos}, the understanding of spatial structure was often overlooked. To better learn the spatio-temporal characteristics and map them to a low-dimensional, expressive latent space, we propose a motion embedding module named Body-Part Attention-based Spatio-Temporal VQ-VAE.

\textbf{Body-part Attention-based Spatio-Temporal feature extraction}\quad
Although the human body has many joints, the body part should be the basic unit that best represents the semantic information of motion. The interaction between joints within the same body part ultimately combines into the motion of the entire human. Some previous works have already realized this point~\cite{jang2022motion,aberman2020skeleton,hu2023pose}. Based on this observation, we propose a spatio-temporal encoder based on body-part attention to extract spatio-temporal features.

\begin{figure}[ht]
    \centering
    \includegraphics[width = 0.48\textwidth]{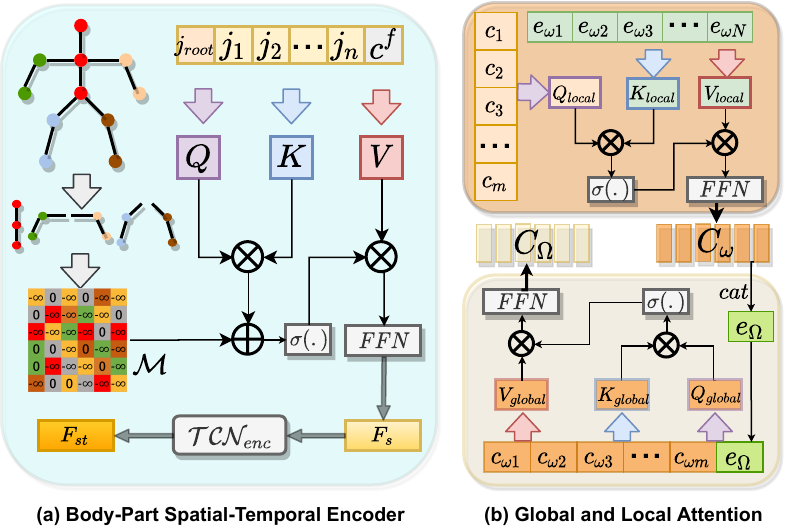}
    \caption{\textbf{BPST and GLA module}. (a)We divide the skeleton into five body parts and use the spatial transformer to learn the body-part attention to extract the spatial feature$F_s$. The final $F_{st}$ is obtained by $TCN_{enc}$. (b)To better learn the cross-modal relationship between text and motion, we extract the word-level and sentence-level text features($C_\omega$ and $C_\Omega$) and use them to calculate the local motion-word cross-attention and the global motion-sentence conditional self-attention, respectively.}
    \label{img:03}
    \vspace{-10pt}
\end{figure}

As shown in Figure~\ref{img:03}(a), we divide the human body with $n$ joints into five body parts: \{Torso, Left Arm, Right Arm, Left Leg, Right Leg\}, each containing its own set of joints. To apply transformer in the spatial dimension, we rearrange the representation of T2M~\cite{guo2022generating} to gather information of each joint, leading to $n+1$ motion tokens of each frame: $\bar{x}_i=\{j_{root},j_1,j_2,...,j_{n-1},c^f\}$. Next, we propose a spatial transformer $ {Trans_{enc}}$ based on body-part attention. Before computing self-attention, we map all tokens to the same dimension through different linear mapping:
\begin{equation}
\begin{aligned}
&Q^{\mathcal{I}}=W_q^{\mathcal{I}}\mathcal{J},K^{\mathcal{I}}=W_k^{\mathcal{I}}\mathcal{J},V^{\mathcal{I}}=W_v^{\mathcal{I}}\mathcal{J}
\end{aligned}
\label{euqa:01}
\end{equation}
where the superscript $\mathcal{I}=\{root,others,contact\}$ of $Q$, $K$, $V$ and $W_q$, $W_k$, $W_v$ is a denotes set representing the root joint, other joints, and foot contact. And $\mathcal{J}=\{j_{root}, j_i,c_f\}$ are the motion tokens. Finally, we concatenate the mapping results of different tokens to obtain the final $Q$, $K$,$V$. When calculating body-part attention, we define an adjacency mask $\mathcal{M} = \{m_{i,j}\} \in \mathbb{R}^{(n+1) \times (n+1)}$ according to body part division. The dimension of $\mathcal{M}$ becomes $n+1$ because foot contact $c_f$ is treated as an extra joint to avoid foot slide. If joint $i$ and $j$ are in the same body part, $m_{i,j}=0$, otherwise $-\infty$we add the adjacency relation mask $\mathcal{M}$. The body-part attention is calculated by:

\begin{equation}
BPAtt = softmax(\frac{QK^T \oplus \mathcal{M}}{\sqrt{D}})V
\label{euqa:02}
\end{equation}

Finally, we repeat the above steps to calculate multi-head self-attention and obtain the final spatial feature $ {f_s}$ through a feed-forward network($FFN$). $\mathcal{M}$ limits the self-attention calculation within the body part. After extracting the spatial features, we use a temporal encoder $ {TCN_{enc}}$ consisting of multiple temporal convolutional layers to obtain the motion spatio-temporal feature $f_{st}$.

\textbf{Latent space learning based on VQ-VAE}\quad
Inspired by T2M-GPT~\cite{zhang2023t2m}, we use VQ-VAE to learn the latent motion space. First, we construct a learnable codebook $C = \{c_i\} \in \mathbb{R}^{K*D_C}$ as a set of discrete latent motion representations. Then, we map the motion sequence $X$ to latent features $F_{st}$ through our spatio-temporal feature extraction module and then quantize it to a discrete latent representation. We use exponential moving average (EMA) and codebook reset~\cite{williams2020hierarchical} to train VQ-VAE and optimize it with the following loss function:
\begin{equation}
\mathcal{L}= \mathcal{L}{recon}+\alpha \mathcal{L}{vel}+ \mathcal{L}{emb}+ \beta \mathcal{L}{com}
\label{equa:04}
\end{equation}
where:
\begin{equation}
\begin{aligned}
&\mathcal{L}{rec} =\mathcal{L}1(X,\hat{X}),\quad \mathcal{L}{vel} =\mathcal{L}1(V,\hat{V})\\
&\mathcal{L}{emb} =\left\lVert sg[F_{st}]-C \right \rVert_2, \mathcal{L}{com} =\left\lVert\ F_{st}-sg[C]\right\rVert_2
\end{aligned}
\label{equa:05}
\end{equation}

$V$ is velocity and $\hat{\quad }$ indicates reconstructed. $\mathcal{L}{rec}$, $\mathcal{L}{vel}$, $\mathcal{L}{emb}$ and $\mathcal{L}{com}$ are the reconstruction term, smooth velocity term, embedding term and commit term, respectively.

\subsection{The Text-driven Motion Generation Module}
We observe that there are two-level correspondences in the cross-modal relationship between motion and text: (1) the correspondence between words in text and sub-segments in motion sequences. For example, in the sentence "With both arms outstretched at the side, bent at elbow, the person raises arms straight over head", words such as "outstretched", "bent", "raises" and "arm" are essential. If we do not consider this correspondence, important words may be ignored, resulting in generated motion inconsistent with the given text; (2) the correspondence between the overall semantics of the text and the whole motion sequence. The temporal properties of the motion are continuous, but the textual ones may not be. In the above sentence, "outstretched" and "bent" has a sequential order, while the actions they represent occur simultaneously. Thus the overall correspondence between the sentence and motion also plays an indispensable role in text-driven motion generation. Previous work usually only considered one aspect of the correspondences~\cite{zhang2022motiondiffuse,tevet2023human,petrovich2022temos,zhang2023t2m,chen2023mld} or simply concatenate two of them~\cite{guo2022generating}. To achieve better text-driven motion generation, we propose a Global and Local Attention Generative Transformer to simultaneously consider these two correspondences, using different levels of attention to better learn the cross-modal relationship between text and motion (as shown in Figure~\ref{img:03}(b)).

\textbf{Local cross-modal relationship}\quad
For a text description $\Omega$ with $N$ words and the corresponding motion discrete representations $C=\{c_i\}_{i=1}^m$, we first extract word-level features $e_\omega = \{e_{\omega_i}\}_{i=1}^N$ using the pre-trained model CLIP~\cite{radford2021learning}. To learn the importance of each word in the cross-modal correspondence, we propose a local cross-modal transformer $ {Trans_{local}}$ to calculate the cross-attention between $e_\omega$ and $C$. Specifically, we use $e_\omega$ as the key and value and $C$ as the query:
\begin{equation}
Q_{local}=W_q C,K_{local}=W_ke_\omega,V_{local}=W_ve_\omega
\label{euqa:06}
\end{equation}
Then, we compute their cross-attention:
\begin{equation}
CrossAtt = softmax(\frac{Q_{local}K_{local}^T }{\sqrt{D}})V_{local}
\label{euqa:07}
\end{equation}

In this way, we can estimate the importance of each word in the sentence for motion generation(seeing Fig.~\ref{img:07}) and learn the cross-modal relationship at the local level. Finally, we input $CrossAtt$ into a $FFN$ to obtain the joint motion-words embedding $C_\omega=\{c_{\omega i}\}_{i=1}^m$.

\textbf{Global cross-modal relationship}\quad
Similar to extracting word-level features, we also use a CLIP to extract sentence-level features $e_\Omega \in \mathbb{R}^{1 \times D}$ from the text. Since the size of $e_\Omega$ is $1 \times D$, performing a softmax weighted sum on it is meaningless. To better learn global cross-modal relationship, we adopt the idea of T2M-GPT~\cite{zhang2023t2m} and use a transformer based on conditional self-attention, called $ {Trans_{global}}$. Specifically, we concatenate $e_\Omega$ with $C_\omega$ on temporal dimension to calculate self-attention between $e_\Omega$ and $C_\omega$. During this process, $e_\Omega$ is fully involved in the attention calculation with $C_\omega$, which is beneficial for learning the global cross-modal relationship. 

To obtain more diverse generation results, we input $C_\Omega$ into a generative transformer $ {TCN_{gen}}$ to predict the probability $p(c_{m+1}|C_{\Omega})$ of the next motion code in the codebook, and optimize the maximum likelihood:

\begin{equation}
\mathcal{L} = -\mathbb{E}_{c_{m+1} \sim p(C)}[log p(c_{m+1}|C_{\Omega})]
\label{euqa:05_15}
\end{equation}

During testing, we directly input the sentence-level feature $e_\Omega$ into $ {TCN_{global}}$ to predict the first motion code $c_1$, then continuously generate a motion code sequence $C_{gen}$. Finally, we input it into the motion decoder $ {TCN_{enc}}$ to obtain the human motion sequence $\hat{X}=\{\hat{x}_i\}_{i=1}^T$.

\begin{table*}[ht]

\begin{center}
\begin{tabular}
{p{2.15cm}p{1.70cm}p{1.70cm}p{1.70cm}p{1.70cm}p{1.70cm}p{1.70cm}p{1.70cm}}
\hline
\multirow{2}{*}{Methods}&\multicolumn{3}{c}{R Precision$\uparrow$}&\multirow{2}{*}{FID.$\downarrow$}&\multirow{2}{*}{MM-D.$\downarrow$}& \multirow{2}{*}{Div.$\rightarrow$}& \multirow{2}{*}{MM.$\uparrow$}\\ \cline{2-4}
& Top-1& Top-2& Top-3&&&&\\
\hline

 Real& 0.511 \scriptsize $\pm$ .003& 0.703  \scriptsize $\pm$ .003& 0.797  \scriptsize $\pm$ .002& 0.002  \scriptsize $\pm$ .000&2.974  \scriptsize $\pm$ .008 &9.503 \scriptsize $\pm$ .065 &-\\
 \hline

Hier~\cite{ghosh2021synthesis} & 0.301 \scriptsize $\pm$ .002& 0.425  \scriptsize $\pm$ .002& 0.552  \scriptsize $\pm$ .004& 6.523  \scriptsize $\pm$ .024&5.012  \scriptsize $\pm$ .018 &8.332 \scriptsize $\pm$ .042 &-\\

TM2T~\cite{guo2022tm2t} & 0.424 \scriptsize $\pm$ .003& 0.618  \scriptsize $\pm$ .003& 0.729  \scriptsize $\pm$ .002& 1.501  \scriptsize $\pm$ .017&3.467  \scriptsize $\pm$ .011 &8.589 \scriptsize $\pm$ .076 &2.424 \scriptsize $\pm$ .093\\

T2M~\cite{guo2022generating} & 0.455 \scriptsize $\pm$ .003& 0.636  \scriptsize $\pm$ .003& 0.736  \scriptsize $\pm$ .002& 1.087  \scriptsize $\pm$ .021&3.347  \scriptsize $\pm$ .008 &9.175 \scriptsize $\pm$ .083 &2.219 \scriptsize $\pm$ .074\\

MDM~\cite{tevet2023human} & 0.320 \scriptsize $\pm$ .005& 0.498  \scriptsize $\pm$ .004& 0.611  \scriptsize $\pm$ .007& 0.544  \scriptsize $\pm$ .044&5.566 \scriptsize $\pm$ .027 &\textbf{9.559} \scriptsize $\pm$ .086 &\textbf{2.799} \scriptsize $\pm$ .072\\

MD~\cite{zhang2022motiondiffuse} & \underline{0.491} \scriptsize $\pm$ .001& \underline{0.681}  \scriptsize $\pm$ .001& \underline{0.782}  \scriptsize $\pm$ .001& 0.630  \scriptsize $\pm$ .001&\underline{3.113}  \scriptsize $\pm$ .001 &\underline{9.410} \scriptsize $\pm$ .049 &1.553 \scriptsize $\pm$ .042\\

MLD~\cite{chen2023mld} & 0.481 \scriptsize $\pm$ .003& 0.673 \scriptsize $\pm$ .003& 0.772  \scriptsize $\pm$ .002& 0.473  \scriptsize $\pm$ .013&3.196  \scriptsize $\pm$ .010 &9.724 \scriptsize $\pm$ .082 &2.413 \scriptsize $\pm$ .079\\ 

T2M-GPT~\cite{zhang2023t2m} & \underline{0.491} \scriptsize $\pm$ .003& 0.680  \scriptsize $\pm$ .003& 0.775  \scriptsize $\pm$ .002& \underline{0.116}  \scriptsize $\pm$ .004&3.118  \scriptsize $\pm$ .011 &9.761 \scriptsize $\pm$ .081 &1.856 \scriptsize $\pm$ .011\\

\hline

Ours & \textbf{0.499} \scriptsize $\pm$ .003& \textbf{0.690}  \scriptsize $\pm$ .002& \textbf{0.786}  \scriptsize $\pm$ .002& \textbf{0.112}  \scriptsize $\pm$ .006&\textbf{3.038}  \scriptsize $\pm$ .007 &9.700 \scriptsize $\pm$ .090 &\underline{2.452} \scriptsize $\pm$ .051\\
\hline
\end{tabular}
\end{center}
\caption{\textbf{Quantitative evaluation on the testset of HumanML3D.} We report the metrics following T2M~\cite{guo2022generating} and repeat 20 times to get the average results with 95\% confidence interval. The $\downarrow, \uparrow, and \rightarrow$ denote the lower, higher, and closer to Real are better, respectively. The best results are marked in bold and the second best is underlined. Our method achieves significant improvement on almost all metrics.}
\label{table:01}
\end{table*}

\begin{table*}[ht]

\begin{center}
\begin{tabular}
{p{2.15cm}p{1.70cm}p{1.70cm}p{1.70cm}p{1.70cm}p{1.70cm}p{1.70cm}p{1.70cm}}
\hline
\multirow{2}{*}{Methods}&\multicolumn{3}{c}{R Precision$\uparrow$}& \multirow{2}{*}{FID.$\downarrow$}&\multirow{2}{*}{MM-D.$\downarrow$}& \multirow{2}{*}{Div.$\rightarrow$}& \multirow{2}{*}{MM.$\uparrow$}\\ \cline{2-4}
&Top-1&Top-2&Top-3&&&&\\
\hline

 Real& 0.424 \tiny $\pm$ .005& 0.649  \tiny $\pm$ .006& 0.779  \tiny $\pm$ .006& 0.031  \tiny $\pm$ .004&2.788  \tiny $\pm$ .012 &11.08 \tiny $\pm$ .097 &-\\
 \hline

Hier~\cite{ghosh2021synthesis} & 0.255 \tiny $\pm$ .006& 0.432  \tiny $\pm$ .007& 0.531  \tiny $\pm$ .007& 5.203  \tiny $\pm$ .107&4.986 \tiny $\pm$ .027 &9.563 \tiny $\pm$ .072 &-\\

TM2T~\cite{guo2022tm2t} & 0.280 \tiny $\pm$ .005& 0.463  \tiny $\pm$ .006& 0.587 \tiny $\pm$ .005& 3.599  \tiny $\pm$ .153&4.591  \tiny $\pm$ .026 &9.473 \tiny $\pm$ .117 &\textbf{3.292} \tiny $\pm$ .081\\

T2M~\cite{guo2022generating} & 0.361 \tiny $\pm$ .006& 0.559  \tiny $\pm$ .007& 0.681 \tiny $\pm$ .007& 3.022 \tiny $\pm$ .107&3.488  \tiny $\pm$ .028 &10.72 \tiny $\pm$ .145 &2.052 \tiny $\pm$ .107\\

MDM~\cite{tevet2023human} & 0.164 \tiny $\pm$ .004& 0.291  \tiny $\pm$ .004& 0.396  \tiny $\pm$ .004& \underline{0.497}  \tiny $\pm$ .021&9.191 \tiny $\pm$ .022 &10.85 \tiny $\pm$ .109 &1.907 \tiny $\pm$ .214\\

MD~\cite{zhang2022motiondiffuse} & \textbf{0.417} \tiny $\pm$ .004& \underline{0.621}  \tiny $\pm$ .004& \underline{0.739}  \tiny $\pm$ .004& 1.954 \tiny $\pm$ .064&\textbf{2.958} \tiny $\pm$ .005 &\textbf{11.10} \tiny $\pm$ .143 & 0.730\tiny \ $\pm$ .013\\

MLD~\cite{chen2023mld} & 0.390 \tiny $\pm$ .008& 0.609 \tiny $\pm$ .008& 0.734  \tiny $\pm$ .007& \textbf{0.404}  \tiny $\pm$ .027&3.204  \tiny $\pm$ .027&10.80 \tiny $\pm$ .117 &2.192 \tiny $\pm$ .071\\

T2M-GPT~\cite{zhang2023t2m}& 0.402 \tiny $\pm$ .006& 0.619  \tiny $\pm$ .005& 0.737  \tiny $\pm$ .006& 0.717 \tiny $\pm$ .041&3.053  \tiny $\pm$ .026 &10.86 \tiny $\pm$ .094 &1.912 \tiny $\pm$ .036\\

\hline

Ours& \underline{0.413} \tiny $\pm$ .006& \textbf{0.632}  \tiny $\pm$ .006& \textbf{0.751}  \tiny $\pm$ .006& 0.870  \tiny $\pm$ .039&\underline{3.039}  \tiny $\pm$ .021 &\underline{10.96} \tiny $\pm$ .123 &\underline{2.281} \tiny $\pm$ .047\\
\hline
\end{tabular}
\end{center}
\caption{\textbf{Quantitative evaluation on the testset of KIT-ML.} The experimental settings are the same as Table~\ref{table:01}.We report the metrics following T2M~\cite{guo2022generating} and repeat 20 times to get the average results with 95\% confidence interval. The best results are marked in bold and the second best is underlined.}
\label{table:02}
\end{table*}

\section{Experiments}
In this section, we first introduce the \textbf{dataset and evaluation metrics} in Sec.\ref{ssec:dataset}, \textbf{the implementation details} are demonstrated in supplementary materials. \textbf{Quantitative and qualitative evaluations} will be presented in Sec.\ref{ssec:sota_quanti} and~\ref{ssec:sota_quali}, respectively. Then, we analyze the main components of the method through \textbf{ablation study} in Sec.\ref{ssec:ablation_study}. Finally, we further discuss our \textbf{attention mechanism} in Sec.\ref{ssec:att}. Due to space limitations, we demonstrate our \textbf{failure cases} and \textbf{limitations} in the supplementary material.

\subsection{Dataset and Metrics}\label{ssec:dataset}
\textbf{KIT Motion-Language~\cite{plappert2016kit}} is the first 3D human motion dataset with text labels, consisting of a subset of KIT~\cite{Mandery2015a} and CMU~\cite{cmu} datasets. It contains 3911 motion sequences and 6353 sequence-level text annotations, with an average of 9.5 words per annotation. 

\textbf{HumanML3D~\cite{guo2022generating}:} HumanML3D added detailed text labels to the motion in AMASS dataset, creating a motion-language dataset. It contains 14616 motion segments with a total duration of 28.59 hours. The average length of per segment is 7.1 seconds, and the longest and shortest motions are 10 and 2 seconds, respectively. The dataset contains 44970 text descriptions with an average length of 12 words, covering 5317 unique words. We follow the split of T2M~\cite{guo2022generating} for our training, validation, and testing on both KIT-ML and HumanML3D.

\textbf{Evaluation Metrics}\quad
We adopt the evaluation metrics following T2M~\cite{guo2022generating}, including Frechet Inception Distance(FID.), R-precision, Multi-Modal distance(MM-D.), Diversity(Div.) and Multi-Modality(MM.), which are evaluating the distribution distance between generated motion and ground-truth, the consistency of text and generated motion, the euclidean distance of motion feature and text feature, the diversity of whole generated motion and the diversity of generated motion from the same text, respectively.

\begin{figure*}[t]
    \centering
    \includegraphics[width = 0.98\textwidth]{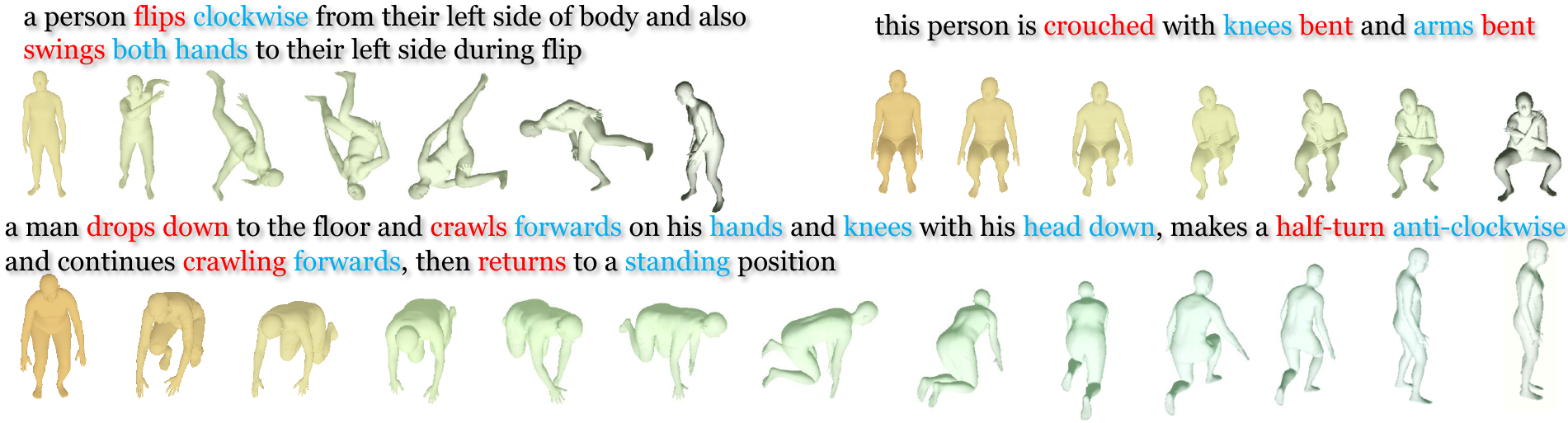}
    \caption{\textbf{More visualization results on HumanML3D}. Our method achieves high-quality motion, including complex actions like "flip", "crouch", and "crawl", which are realistic and highly consistent with the given text. The change of color indicates the passage of time.}
    \label{img:04}
\end{figure*}

\subsection{Quantitative Evaluation}\label{ssec:sota_quanti}

We use the metrics mentioned above to evaluate the quantitative effects of our method and compare them with recent state-of-the-art works, including Hier~\cite{ghosh2021synthesis}, TM2T~\cite{guo2022tm2t}, T2M~\cite{guo2022generating}, MDM~\cite{tevet2023human}, MotionDiffuse(MD)~\cite{zhang2022motiondiffuse}, MLD~\cite{chen2023mld}, and T2M-GPT~\cite{zhang2023t2m}. The results of these methods were obtained from their pre-trained models or related papers.

\textbf{HumanML3D:} As shown in Table~\ref{table:01}, our method have significant improvements in almost all evaluation metrics compared to previous works. The performance of text-driven motion generation has two aspects: the quality of the generated motion itself and its consistency with the given text description. In terms of text consistency, our method has a significant improvement in the accuracy of top 1, 2, and 3 compared to previous works, which are also very close to the results of real data. This indicates that the consistency of our generated motion with the text is the highest, which is also evidenced by our method's lowest MM-D. As for the quality of the generated motion itself, our method has the lowest FID of all methods, indicating that the quality of our generated motion is natural, realistic, and close to real data. It is worth noting that Div can only measure the diversity of motion to a certain extent, as Div may also be large when the generated motion has various artifacts. The more compelling diversity metric is MM, of which our method is the second highest.

\textbf{KIT-ML:}
We evaluate the same metrics on KIT-ML. The results are shown in Table~\ref{table:02}. Since MD~\cite{zhang2022motiondiffuse}, MDM~\cite{tevet2023human}, and MLD~\cite{chen2023mld} used ground-truth lengths in their evaluation, it is understandable that they achieved good results whereas our method has better or comparable performance in all metrics(except FID) without the need for ground-truth lengths. And there is a significant improvement over other methods that do not require ground-truth length, like T2M~\cite{guo2022generating} and T2M-GPT~\cite{zhang2023t2m}.

\subsection{Qualitative Evaluation}\label{ssec:sota_quali}
\textbf{Text-driven Motion Generation:}
We present the visualization results of our approach for long text generation with SMPL~\cite{loper2015smpl} model(Figure~\ref{img:01} and ~\ref{img:04}). The figure shows multiple motion results generated by our method, which are high quality and match the textual description very well. The generated motion types include walking on the ground and upstairs, running, crouching, crawling, flipping, and other actions that involve both movements in situ and global displacement, which fully demonstrate the superiority of our method. Moreover, our method can generate diverse sequences for the same input text while ensuring motion consistency. The diversity results and more visualization results are shown in the supplementary materials and demo.
\begin{figure}[t]
    \centering
    \includegraphics[width = 0.48\textwidth]{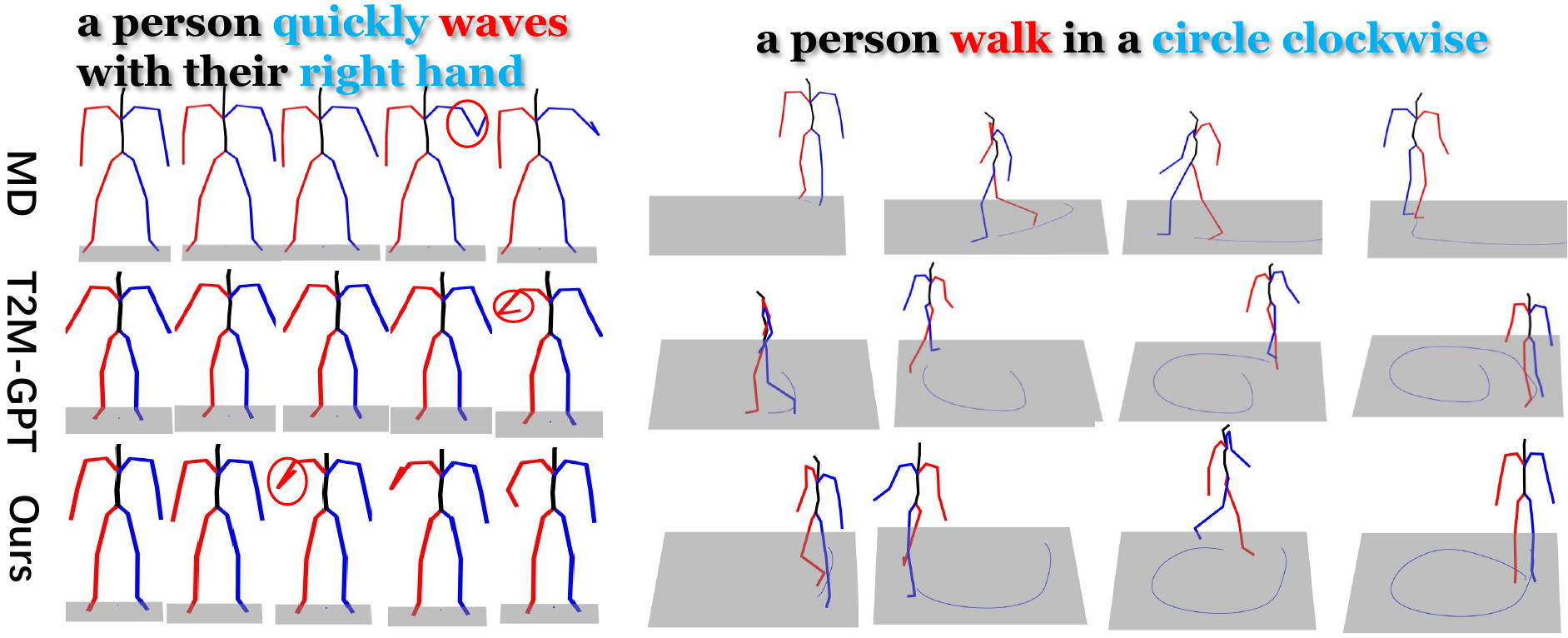}
    \caption{\textbf{Qualitative comparison.} We compare our method qualitatively with MD and T2M-GPT, who achieved the most similar results to us in the quantitative evaluation. The results generated by ours more precisely match the textual descriptions. For a more intuitive comparison, we demonstrate skeleton sequences here.}
    \label{img:05}
\end{figure}

\textbf{Comparison with Previous Works:}
To demonstrate the effectiveness of our method, we conduct qualitative comparison on HumanML3D with MD~\cite{zhang2022motiondiffuse} and T2M-GPT~\cite{zhang2023t2m}, which achieve the most similar results to us in the quantitative evaluation. The results are shown in Figure~\ref{img:05}. The left sequences correspond to the text "a person quickly waves with their right hand", which emphasizes the words "quickly" and "wave right hand". We can see that the motion generated by MD~\cite{zhang2022motiondiffuse} waves the left hand, while the motion generated by T2M-GPT~\cite{zhang2023t2m} ignores the word "quickly" and pauses for a moment before waving. Our result matches the text description best. The sequences on the right correspond to the text "a person walks in a circle clockwise". The motion generated by MD~\cite{zhang2022motiondiffuse} stops after only half a circle, while the trajectory generated by T2M-GPT~\cite{zhang2023t2m} is not a true circle. Our result's trajectory is a more seamless circle. The better performance on text-driven motion generation relies on the learning of the cross-modal relationship between text and motion, which allows our model to focus more attention on essential words.

\begin{figure}[t]
    \centering
    \includegraphics[width = 0.48\textwidth]{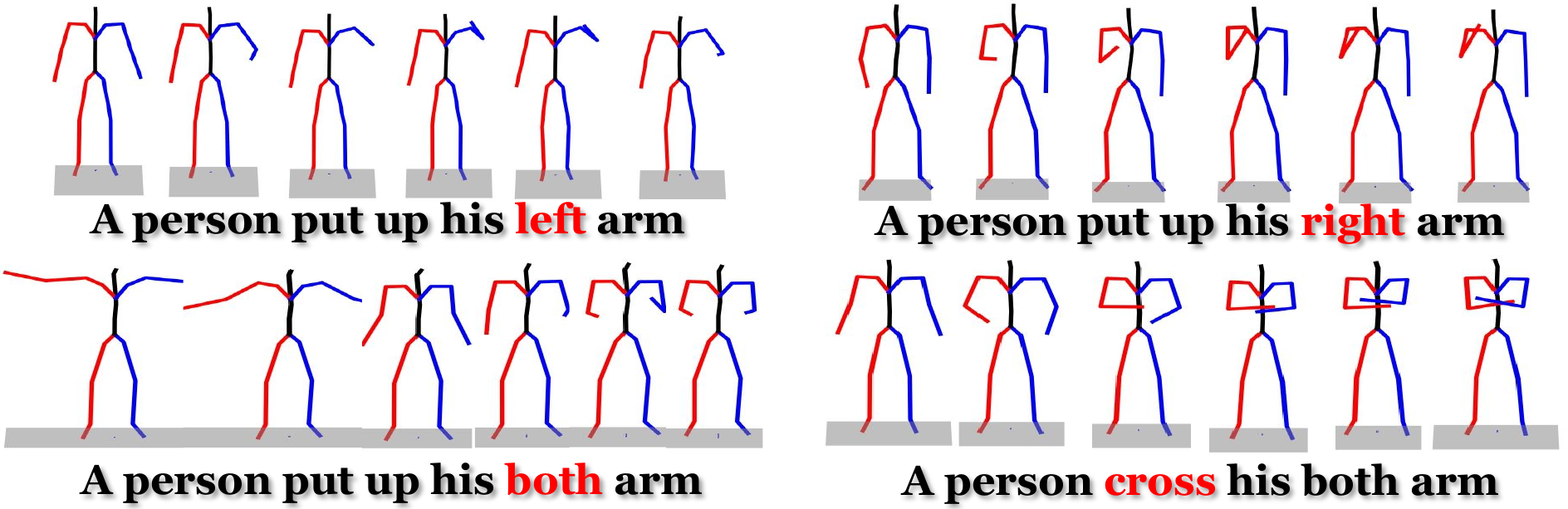}
    \caption{\textbf{Fine-grained generation}. Our method can generate high-quality motion when a single word in a given text is replaced, whether it's a verb or an adjective. For a more intuitive visualization, we demonstrate skeleton sequences here.}
    \label{img:06}
\end{figure}

\begin{figure}[t]
    \centering
    \includegraphics[width = 0.45\textwidth]{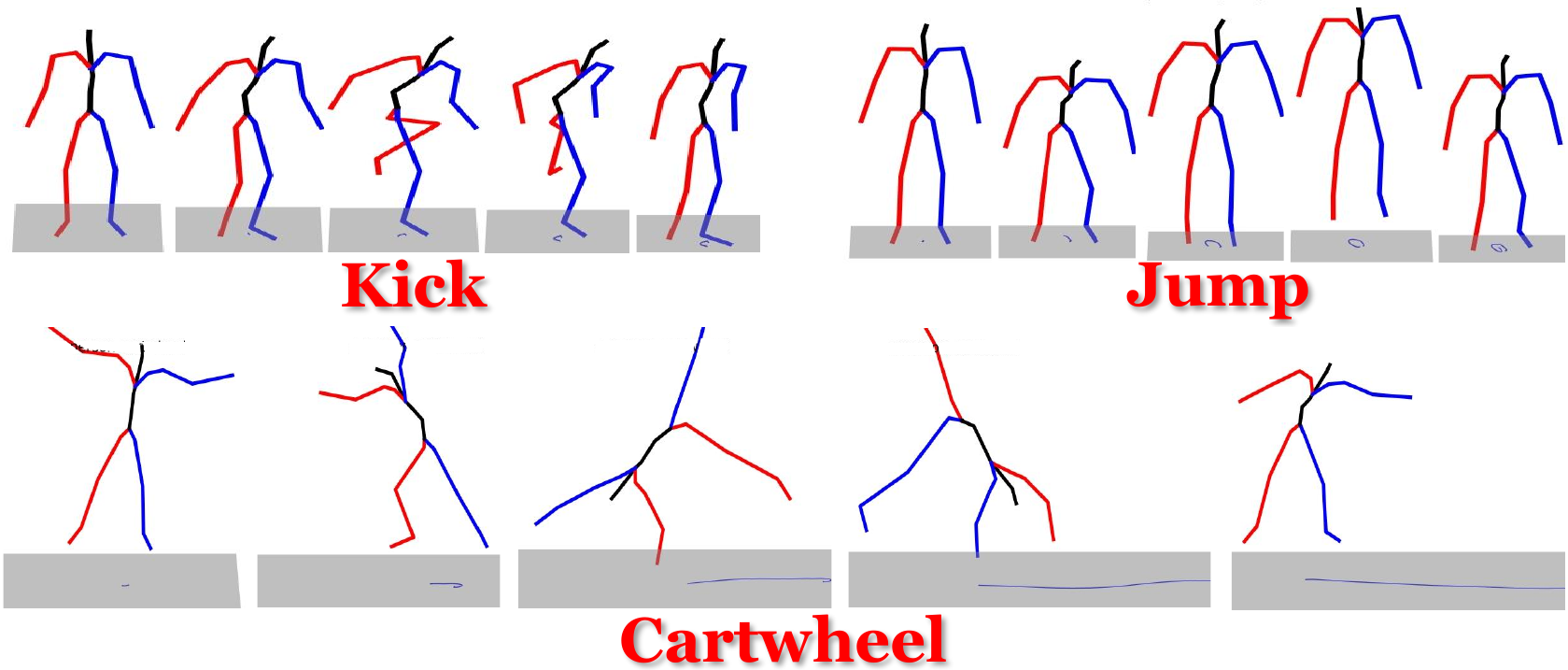}
    \caption{\textbf{Action to motion}. Our method can generate high-quality motion with a given action label expressed by a word or a phrase.}
    \label{img:08}
\end{figure}

\textbf{Fine-grained Motion Generation:}
When only one word is changed in a sentence, the resulting change in the text encoding may be very small, but the corresponding motion sequence can be completely different. Thanks to the body-part attention-based spatial features extraction and the local cross-modal relationship learning, our method can achieve fine-grained motion synthesis by changing a single word in the sentence(Figure~\ref{img:06}). As we modify the word "left" to "right" and "both", the generated motions change from raising the left hand to raising the right and both arms. In addition to replacing adjectives, when we change the verb by replacing "put up" with "cross", the model can also generate the motion of crossing both arms. This demonstrates that our method is very sensitive to small changes in the input text, which is more helpful in keeping the consistency between the generated motion and the text.

\textbf{Action to Motion Generation:}
In addition to long text-driven motion generation, our method can also achieve high quality action to motion generation without any extra training, as shown in Figure~\ref{img:08}. With only a single word or phrase representing the action as input, our method can still generate motion with high quality.

\subsection{Ablation Study}\label{ssec:ablation_study}
We conducted an ablation study to investigate the effect of each module in our method on two aspects: the quality of the generated motion itself and its consistency with the given text description. 

\textbf{Effect of BPST:}
The high-quality low-dimensional motion representation can simplify motion modeling, making the generated motion more realistic and helpful for cross-modal learning between text and motion. To demonstrate the effect of our proposed BPST module, we remove the BPST and directly input the motion representation into the temporal convolutional layer to learn the VQ-VAE. The AS4 in Table~\ref{table:03} shows the results of this ablation study. We can see that the three R-Precision metrics decrease to some extent compared to the complete version(AS6), while FID increases significantly. R-Precision measures the consistency between the generated motion and text description, while FID reflects both the quality of the generated motion itself and the consistency between text and motion. The experimental results indicate that our BPST module significantly impacts the quality of generated motion while also improving the consistency between text and motion to some extent. In addition, we further demonstrated the effectiveness of BPST by comparing the reconstruction error and commitment error of VQ-VAE with T2M-GPT(Table.~\ref{table:vae}).

\begin{table}[t]
\scriptsize
\renewcommand\arraystretch{1.1}
\begin{center}
\begin{tabular}
{p{0.4cm}p{0.3cm}p{0.35cm}p{1.15cm}p{1.15cm}p{1.15cm}p{1.15cm}}
\hline
\multirow{2}{*}{\footnotesize BPST}&\multirow{2}{*}{\footnotesize GA}&\multirow{2}{*}{\footnotesize LA}&\multicolumn{3}{c}{\footnotesize R Precision$\uparrow$}&\multirow{2}{*}{\quad \footnotesize FID.$\downarrow$}\\ \cline{4-6}
&&&\quad \footnotesize Top-1&\quad \footnotesize Top-2&\quad \footnotesize Top-3&\\
\hline

&&& 0.472 \tiny $\pm$ .002& 0.665  \tiny $\pm$ .002& 0.759  \tiny $\pm$ .002& 0.359  \tiny $\pm$ .007\\

\ \Checkmark&\ \Checkmark&& 0.484 \tiny $\pm$ .002& 0.670  \tiny $\pm$ .003& 0.769  \tiny $\pm$ .002& 0.183  \tiny $\pm$ .005\\

\ \Checkmark &&\ \Checkmark& 0.493\tiny $\pm$ .002&0.681\tiny $\pm$ .002&0.779\tiny $\pm$ .002&0.156\tiny $\pm$ .004\\

&\ \Checkmark&\ \Checkmark& \underline{0.495} \tiny $\pm$ .002& \underline{0.682}  \tiny $\pm$ .002& \underline{0.781}  \tiny $\pm$ .002& 0.213  \tiny $\pm$ .004\\

\ \Checkmark&\ \Checkmark&\ \Checkmark& \textbf{0.499} \tiny $\pm$ .003& \textbf{0.690}  \tiny $\pm$ .002& \textbf{0.786}  \tiny $\pm$ .002& \textbf{0.112}  \tiny $\pm$ .006\\
\hline

\end{tabular}
\end{center}
\caption{\textbf{Ablation study}. Ablation studies are conducted on HumanML3D to verify the effectiveness of body-part attention(BPST), local attention(LA), and global attention(GA). The results are labeled from top to bottom as AS1-AS5. The best results are marked in bold and the second best is underlined.}
\label{table:03}
\end{table}

\begin{table}[t]
\begin{center}
\begin{tabular}
{c|cc}
\hline
&Recon. Error$\downarrow$&Commit Error$\downarrow$\\
\hline

T2M-GPT(w/o BPST)&0.6512&2.1943\\
\hline
Ours(w/ BPST)&\textbf{0.6427}&\textbf{2.0701}\\
\hline
\end{tabular}
\end{center}
\vspace{-7pt}
\caption{Comparison on reconstruction and commitment error of our VQ-VAE and T2M-GPT's.}
\label{table:vae}
\end{table}

\textbf{Effect of GLA:}
The role of the GLA module is to learn the cross-modal relationship between text and motion, ensuring consistency between the generated motion and the given text. To demonstrate the effectiveness of our GLA, we keep the settings of stage 1 unchanged and remove ${Trans_{local}}$ and ${Trans_{global}}$ in turn. The AS2 and AS3 in Table~\ref{table:03} shows the results of this ablation study. We can see that compared to AS5, the R Precision of AS2 and AS3 decreases significantly, and FID decreases slightly. This indicates that GLA can significantly improve the realism of motion and consistency with the text description. In addition, we found that the performance of AS2 is worse than that of AS3, which indicates that LA plays a more important role in text-driven motion generation than GA.

Finally, we remove both BPST and GLA, and use vanilla VQ-VAE and concatenation of motion and text in vanilla transformer. The result is shown in AS1 of Table~\ref{table:03}. All metrics became much worse, indicating that our method is reasonable and effective.

\begin{figure}[t]
    \centering
    \includegraphics[width = 0.48\textwidth]{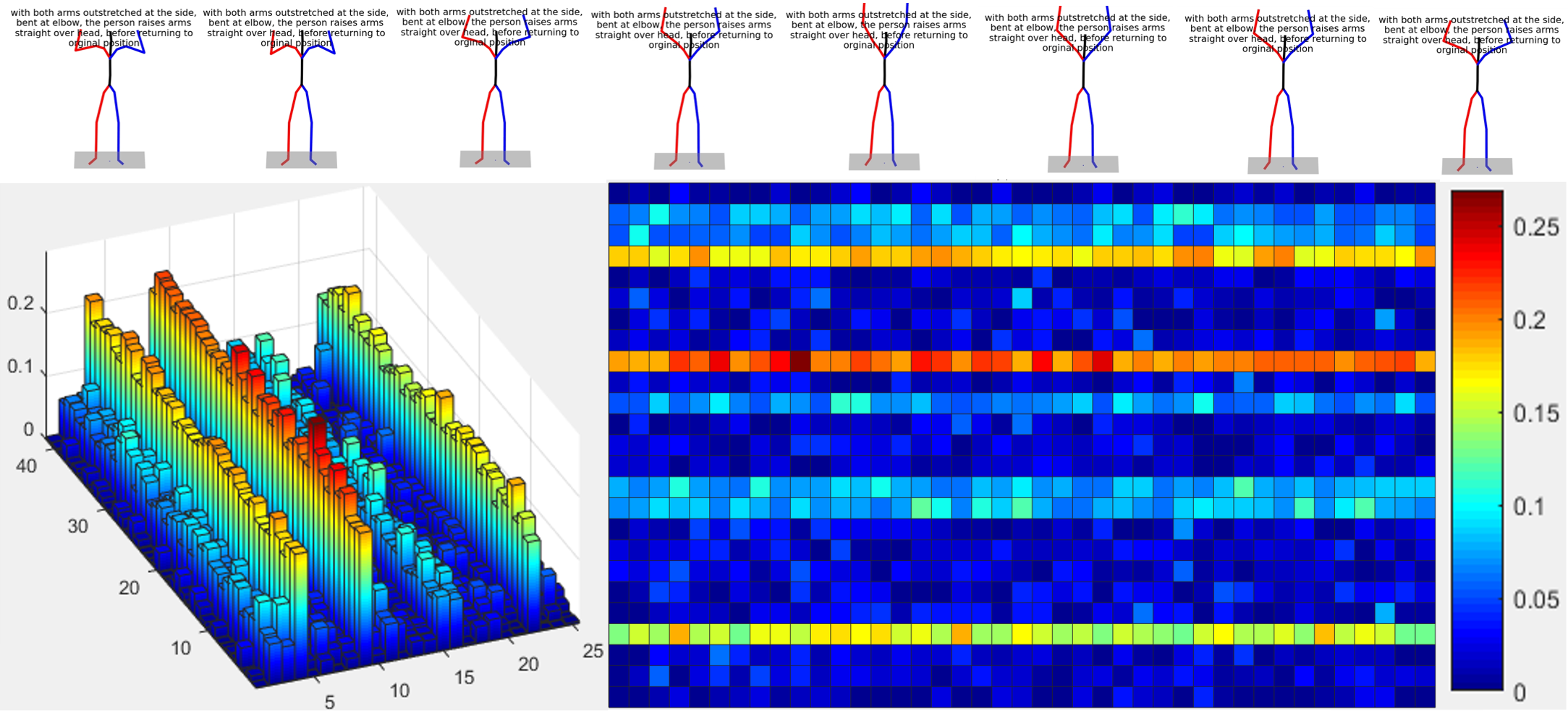}
    \caption{\textbf{Motion-word cross-attention visualization}. We visualize the motion-word cross-attention in 2D and 3D heatmaps to demonstrate that our method does learn the different importance of words in the sentence in the process of text-driven generation.}
    \label{img:07}
\end{figure}

\subsection{Attention Visualization}\label{ssec:att}

To further explain the effectiveness and rationality of our method, we visualize the cross-attention between words and motion. We selecte a generated motion sequence of length 41 described by the text "With both arms outstretched at the side, bent at the elbow, the person raises arms straight overhead, before returning to the original position." The visualization of attention is shown in Figure~\ref{img:07} using 2D and 3D heatmaps. Rows 4, 15, and 22 of the 2D heatmap correspond to the words "outstretched," "raise," and "returning," which represent the main actions of this motion sequence and have the highest attention weights. Rows 2, 3, 9, 11, and 16 correspond to words such as "both", "arms", "bent", "elbow", and "arms", which are related to the body parts and states of them and relevant to the motion sequence. Other words and punctuation in the sentence have weak relationships with the motion and therefore have correspondingly lower weights. This visualization demonstrates that our method can indeed learn local cross-modal attention for each word, which is helpful for learning cross-modal relationships between text and motion. The body-part attention is shown in the supplementary materials.

\subsection{User Perceptual Study}
\begin{figure}[t]
\centering
\includegraphics[width = 0.48\textwidth]{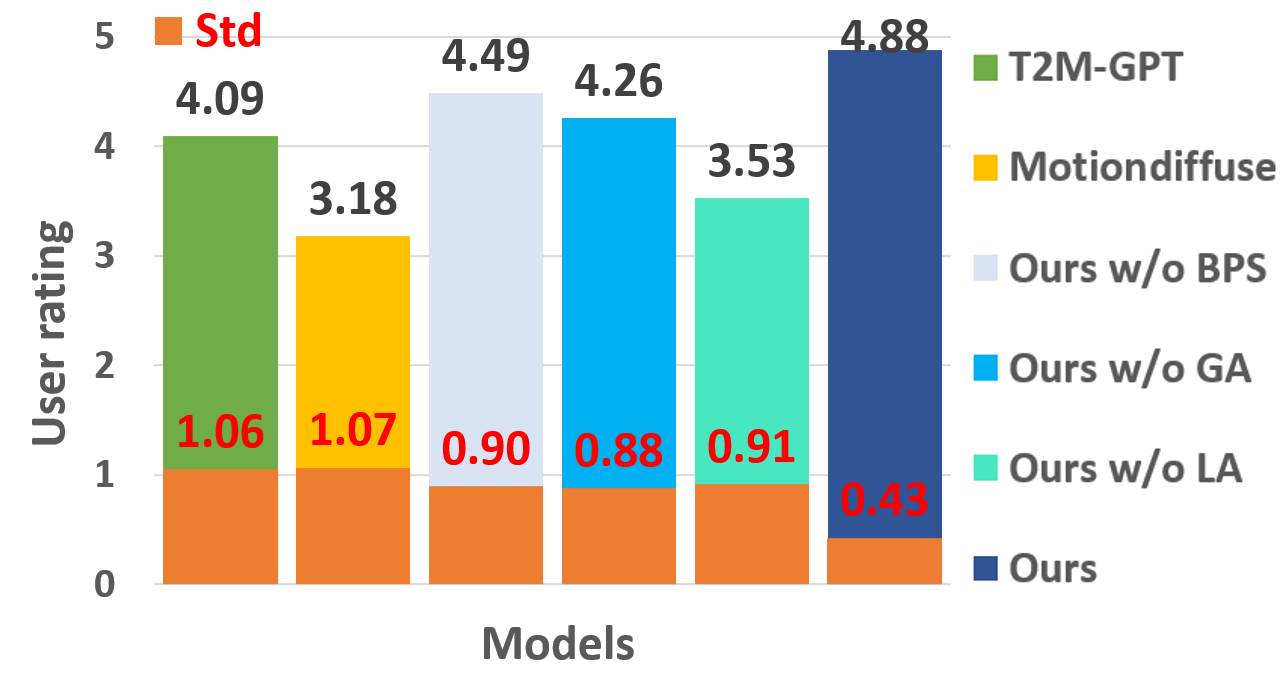}
\caption{User perceptual study. The motion generated by our method is the one that best matches the textual description from the user's perspective.} 
\label{fig:userstudy}
\end{figure}

To illustrate the visual superiority of our approach, we did a user perceptual study, and the results are shown in Fig.~\ref{fig:userstudy}. Specifically, we randomly select 30 texts in testset of HumanML3D and set up 10 manual texts to generate motion using the following 6 models: T2M-GPT, Motiondiffuse, Ours w/o BPS, Ours w/o GA, Ours w/o LA, Ours. We have 30 participants and ask them to rate the generated motion on a scale of 1-6(6 is the best). The evaluation criteria is a combination of motion realism and text-motion consistency. The highest score and lowest std indicate that our method works better and is more robust.

\section{Limitations and Failure Cases}
Despite achieving the state-of-the-art performance, our method still has some limitations and corresponding failure cases as follows(See the supplemental video for the visualization results).

\textbf{(1) Insufficient diversity in long and detailed text-driven generation:} Our method can generate motions with diversity. However, for very long texts with very detailed descriptions in the test set, the diversity of the motions generated by our method is insufficient. We observe that there are usually few motion sequences corresponding to long and detailed text in the HumanML3D dataset. When we want to generate multiple motions with them, the generated motions will be very similar. 

\textbf{(2) Fine-grained generation without ground truth or any similar sample in the dataset:} As mentioned in the main paper, our approach enables fine-grained motion generation by changing a word or phrase in a sentence. However, when the motion represented by the changed word does not have ground truth in the dataset, or does not have any similar motion, our method does not yield results that match the text description well. 

\textbf{(3) Out-of-distribution Generation:} When the given text does not lie within the distribution of the dataset, our method is unable to generate motions consistent to the text. But our method still tries very hard to find the motion similar to this text description. 

\section{Conclusion}
We propose AttT2M, a two-stage method with multi-perspective attention mechanism for text-driven motion generation. Specifically, we use the spatial transformer based on body-part attention and temporal convolution to extract spatio-temporal motion features, and map them into a discrete latent space using VQ-VAE. Next, we learn the cross-modal relationship between the latent motion representation and the text utilizing global and local attention, which means calculating sentence-level conditional self-attention and word-level cross-attention. Finally, a generative transformer is trained to perform motion generation. Extensive experiments demonstrate that our method outperforms state-of-the-art methods in both qualitative and quantitative evaluations, and can accomplish fine-grained motion synthesis and action to motion.
\\
\\
\hspace{-0.5cm}\textbf{\large{Acknowledgements}} 
This work was supported by National Key Research and Development Program of China (NO. 2022YFB3303202).

\newpage

{\small
\bibliographystyle{ieee_fullname}
\bibliography{main}
}

\end{document}